\documentclass{article}

\usepackage{microtype}
\usepackage{graphicx}
\usepackage{booktabs}
\usepackage{hyperref}

\usepackage[preprint]{icml2026}
\usepackage{amsmath,amssymb,mathtools}
\usepackage{array}
\usepackage{tabularx}
\usepackage{multirow}
\usepackage{xcolor}
\usepackage{enumitem}
\usepackage{tikz}
\usetikzlibrary{arrows.meta,fit,positioning}
\usepackage[capitalize,noabbrev]{cleveref}

\definecolor{CocoonInk}{HTML}{172321}
\definecolor{CocoonTeal}{HTML}{2E6D68}
\definecolor{CocoonMint}{HTML}{DDEBE7}
\definecolor{CocoonSand}{HTML}{F1ECE2}
\definecolor{CocoonGray}{HTML}{64706D}

\hypersetup{
  colorlinks=true,
  linkcolor=CocoonTeal,
  citecolor=CocoonTeal,
  urlcolor=CocoonTeal,
  pdftitle={Evidence-Grounded Constraint Checking in Construction Documents},
  pdfauthor={Rashid Mushkani; Hugo Berard; Shin Koseki}
}

\newcommand{\pp}{percentage points}
\newcommand{\decision}[1]{\textsc{#1}}
\newcommand{\meets}{\decision{Meets}}
\newcommand{\notmeets}{\decision{Does Not Meet}}
\newcommand{\missing}{\decision{Missing Information}}
\newcommand{\uncertain}{\decision{Uncertain}}
\newcommand{\dataset}{AEC-Bench}
\newcolumntype{L}[1]{>{\raggedright\arraybackslash}p{#1}}
\newcolumntype{C}[1]{>{\centering\arraybackslash}p{#1}}
\newcolumntype{R}[1]{>{\raggedleft\arraybackslash}p{#1}}
\newcolumntype{Y}{>{\raggedright\arraybackslash}X}
\setlist[itemize]{leftmargin=*,topsep=2pt,itemsep=1pt,parsep=0pt}

\tikzset{
  block/.style={draw=CocoonTeal, rounded corners=2pt, line width=.55pt,
    fill=white, align=center, inner sep=4.5pt, font=\scriptsize},
  data/.style={block, fill=CocoonSand},
  method/.style={block, fill=CocoonMint},
  output/.style={block, fill=white},
  flow/.style={-{Latex[length=2mm]}, line width=.55pt, draw=CocoonGray}
}

\newcommand{\SuiteTasks}{160}
\newcommand{\SuiteProjects}{29}
\newcommand{\CoreTasks}{40}
\newcommand{\CoreProjects}{6}
\newcommand{\CoreTrials}{2,400}
\newcommand{\CoreDelta}{10.6}
\newcommand{\CoreCILow}{4.3}
\newcommand{\CoreCIHigh}{18.0}
\newcommand{\CoreP}{.031}
\newcommand{\BreadthTasks}{120}
\newcommand{\BreadthProjects}{23}
\newcommand{\BreadthTrials}{480}
\newcommand{\BreadthDelta}{-4.1}
\newcommand{\BreadthCILow}{-10.2}
\newcommand{\BreadthCIHigh}{1.9}
\newcommand{\BreadthP}{.209}

\icmltitlerunning{Evidence-Grounded Constraint Checking in Construction Documents}

\begin{document}

\twocolumn[
  \icmltitle{Evidence-Grounded Constraint Checking in Construction Documents}

  \begin{icmlauthorlist}
    \icmlauthor{Rashid Mushkani}{cocoon}
    \icmlauthor{Hugo B\'erard}{cocoon}
    \icmlauthor{Shin Koseki}{cocoon}
  \end{icmlauthorlist}

  \icmlaffiliation{cocoon}{Cocoon Lab}
  \icmlcorrespondingauthor{Rashid Mushkani}{rashid@cocoonlab.ai}
  \icmlkeywords{multimodal document understanding, constraint checking, construction documents, evidence grounding, selective prediction}
  \vskip 0.3in
]

\printAffiliationsAndNotice{}

\begin{abstract}
Professional-document review is a constraint-checking problem in which the
answer depends on relations among text, geometry, pages, and document
revisions. We present an evidence-grounded pipeline that normalizes extracted
facts, executes four-state rules deterministically, retains source spans, and
escalates unresolved cases. We evaluate its PDF evidence allocator on
\SuiteTasks{} expert-referenced tasks from \SuiteProjects{} construction
projects, with a repeated four-system test and a disjoint two-system breadth
extension. In the repeated test, reallocating a four-image cap from retrieved
page overviews to one overview plus three overlapping tiles improves
project--family standardized decision accuracy by \CoreDelta{} percentage
points (95\% project-cluster bootstrap CI \CoreCILow{}--\CoreCIHigh{};
exact $p=\CoreP$). The contrast does not persist in the larger breadth block:
Region-RAG changes accuracy by \BreadthDelta{} points (95\% CI
\BreadthCILow{}--\BreadthCIHigh{}; exact $p=\BreadthP$), and the equal-image
sensitivity favors page breadth. Exact finding localization remains low,
false passes remain common, and repeated-run agreement is poorly calibrated.
The central result is therefore not a universal advantage for tiling, but a
measured resolution--breadth tradeoff that requires rule- and evidence-aware
routing before expert review.
\end{abstract}

\section{Introduction}

Reviewing a construction set requires relations that an unordered text stream
does not preserve. A detail callout is meaningful only at the object where its
leader terminates; a sheet reference is valid only if its destination exists;
and a product submittal satisfies a specification only when values agree
across documents. These are typed constraints over text, geometry, document
structure, and revision state. Retrieval can expose the right terms while
still losing the relation needed for a correct professional judgment.

This paper studies a general machine-learning problem: \emph{evidence-grounded
constraint checking in multimodal professional documents}. Construction
documents provide a demanding testbed because pages are large, evidence is
sparse, and an unchecked false pass can be more consequential than a flagged
case. We use \dataset{} \citep{mankodiya2026aecbench} and score predictions
against a typed inventory of its released expert-generated references. The
primary endpoint is a four-way decision---\meets, \notmeets, \missing, or
\uncertain---and the
secondary endpoints include finding precision and recall, exact task
correctness, false-pass and false-fail rates, calibration, and selective risk.

We implement an evidence-grounded constraint checker with four components:
source adapters, typed facts, deterministic rules, and proof traces. The PDF
adapter combines layout-preserving text retrieval with bounded visual regions.
CAD and IFC use the same source-span interface but are not part of the present
empirical data. A deterministic reducer converts extracted fact states into a
decision, and a fail-closed policy identifies missing, uncertain, low-confidence,
or untraced facts for expert review.

Our central test is deliberately narrow and falsifiable: under the same text
retrieval and a four-image cap, does allocating visual capacity to local detail
improve expert-reference decisions relative to spreading that capacity across
retrieved page overviews? The repeated test contains \CoreTasks{} tasks from
\CoreProjects{} projects, four Google/Azure systems, three evidence policies,
and five calls per cell (\CoreTrials{} trials). A breadth extension adds
\BreadthTasks{} tasks from \BreadthProjects{} disjoint projects for two Google
systems. Independent inference is performed at the source-project level.

The contributions are:
\begin{itemize}
  \item a typed constraint-checking architecture with deterministic reduction,
  source-level evidence contracts, and explicit expert escalation;
  \item a machine-readable expert-reference suite of \SuiteTasks{} tasks in
  seven constraint families and \SuiteProjects{} source projects; and
  \item a controlled evidence-allocation experiment where a
  \CoreDelta-point repeated-test gain changes to \BreadthDelta{} points on
  23 disjoint projects, exposing a resolution--breadth tradeoff.
\end{itemize}

\section{Related Work}

\paragraph{Multimodal documents.}
DocVQA established visual question answering over document images
\citep{mathew2021docvqa}; MMLongBench-Doc studies long documents containing
text and graphics \citep{ma2024mmlongbench}; and M3DocRAG and ColPali examine
multimodal retrieval for multi-page reasoning \citep{cho2024m3docrag,
faysse2025colpali}. Layout-aware encoders and OCR-free document models learn
page structure from pixels and text \citep{huang2022layoutlmv3,
kim2022donut}. These settings make retrieval or representation part of the
evaluated system. Our focus is narrower than general document QA: the output
is a typed constraint decision with asymmetric safety errors, explicit source
provenance, and source-project grouped evaluation.

Multimodal retrieval work typically optimizes which pages enter a long-context
answering model. Our intervention addresses a subsequent allocation problem:
given a small image budget and a fixed ranked text set, should the packet
preserve page breadth or spend capacity on local resolution? The distinction
matters for large-format drawings, where a page overview can preserve global
layout while making dimensions, leaders, and callout text unreadable. We hold
retrieval fixed to isolate this resolution--breadth choice.

\paragraph{Construction-document understanding.}
\dataset{} evaluates executable workflows over drawings, specifications, and
submittals \citep{mankodiya2026aecbench}. DrawingVQA evaluates expert-authored
questions over issued construction drawings, including resolution and
text-only ablations \citep{jung2026drawingvqa}. We retain the former's diverse
professional tasks but restrict the analysis to seven families that express a
constraint judgment, and score generated findings against its expert-generated
reference inventory. Earlier automated checking systems compile building rules
against structured BIM representations \citep{eastman2009automatic,
solihin2015classification}. Our architecture retains deterministic execution
and auditable rule states while admitting facts extracted from PDFs as well as
CAD and IFC sources.

Structured BIM checking assumes that relevant entities and properties already
exist in a queryable model. PDF sets invert that premise: the checker must
recover entities and relations from sparse text and graphics before applying a
rule. This motivates our separation between a learned fact-extraction boundary
and a deterministic rule boundary. It also permits the same rule and trace
objects to accept future CAD/IFC facts without treating untested modalities as
part of the present result.

\paragraph{Tool use, abstention, and measurement.}
Tool-using language-model systems combine retrieval, actions, and generation
\citep{yao2023react}; their observed performance therefore depends on the
complete inference pipeline. Selective prediction asks a system to trade
coverage for lower risk \citep{geifman2017selective}, while calibration asks
whether confidence corresponds to correctness \citep{guo2017calibration}. In
professional review, both require validation at the level of independent
projects rather than treating repeated calls as new cases. Our analysis
reports these quantities but does not infer expert agreement, which the source
release does not provide.

Repeated sampling is sometimes used as a proxy for confidence in generative
systems. For constraint checking, identical answers can reflect a stable
mistake, so repetition supplies an empirical class frequency rather than a
guaranteed probability of correctness. We measure its calibration and retained
risk directly, with task repetitions nested inside source projects.

\section{Evidence-Grounded Constraint Checking}
\label{sec:method}

\subsection{Problem formulation}

Let a task provide documents $D$, a constraint rule $r$, and extracted facts
$F=\{f_j\}$. Each fact has a typed predicate, subject, expected and observed
values, state $s_j\in\{\text{verified},\text{violated},\text{missing},
\text{uncertain}\}$, confidence, and one or more source references. A source
reference identifies a PDF page and normalized region, a CAD entity, or an IFC
entity and property path.

Let $V$, $M$, and $U$ denote the presence of a violated, missing, or uncertain
fact. For a conjunctive review rule, the deterministic reducer is
\begin{equation}
g(F)=
\begin{cases}
\mathrm{DNM}, & V,\\
\mathrm{MI}, & \neg V\land M,\\
\mathrm{U}, & \neg V\land\neg M\land U,\\
\mathrm{M}, & \text{otherwise},
\end{cases}
\label{eq:reducer}
\end{equation}
where DNM, MI, U, and M correspond to \notmeets, \missing, \uncertain, and
\meets.
This precedence is intentionally fail-closed: a supported violation is not
overridden by absent evidence, and incomplete evidence cannot become a pass.

\subsection{Source and fact contracts}

The source layer separates document access from rule execution. Every source
reference contains an immutable document identifier and a short excerpt. PDF
references additionally require a one-indexed page and may carry a normalized
bounding box. CAD and IFC references require an entity identifier and may
carry a field or property path. Invalid page indices, degenerate regions, and
unresolved entity references fail schema validation before a rule is run.

An extractor maps source evidence to typed facts
$f_j=(q_j,o_j,e_j,s_j,c_j,Z_j)$: predicate $q_j$, observed value $o_j$,
expected value $e_j$, state $s_j$, confidence $c_j\in[0,1]$, and source set
$Z_j$. The rule declares the fact identifiers it requires. A verified or
violated fact must cite at least one source; a required fact absent from the
extractor output is inserted as \texttt{missing}. Duplicate fact identifiers
are rejected. These invariants make extraction errors visible to the rule
layer instead of silently converting them into successful checks.

\begin{algorithm}[t]
\caption{Fail-closed constraint reduction}
\label{alg:reduction}
\begin{algorithmic}[1]
\REQUIRE required fact IDs $R$, extracted facts $F$, threshold $\tau$
\FOR{each $r\in R$ in rule order}
  \IF{$r\notin F$}
    \STATE insert $(r,\texttt{missing},1.0,\varnothing)$
  \ENDIF
  \STATE validate type, confidence, and source contract
\ENDFOR
\IF{any required state is \texttt{violated}}
  \STATE $d\leftarrow\notmeets$
\ELSIF{any required state is \texttt{missing}}
  \STATE $d\leftarrow\missing$
\ELSIF{any required state is \texttt{uncertain}}
  \STATE $d\leftarrow\uncertain$
\ELSE
  \STATE $d\leftarrow\meets$
\ENDIF
\STATE emit ordered fact steps, sources, and completeness
\STATE escalate if $d$ is unresolved, $\min_j c_j<\tau$, or a source is absent
\STATE \textbf{return} decision $d$, proof trace, escalation reasons
\end{algorithmic}
\end{algorithm}

The reducer is deliberately small: the same facts always yield the same
decision, and every branch can be inspected without another generative pass.
The present implementation evaluates conjunctions of required facts. More
complex professional rules can expose disjunctions or exceptions as typed
predicates upstream; evaluating that compilation is outside the current
experiment.

\subsection{Machine-readable rule families}

Each family definition contains a stable rule identifier, a natural-language
statement for authoring and review, a fact type, and the subject fields needed
to bind a task instance. Reference-resolution rules bind a source region,
target sheet, and target view; title and callout rules bind text to depicted
content or a leader endpoint; index rules bind an index entry to a title
block. Specification and submittal rules bind required and observed values to
clause and drawing sources. The rule files also version the four decision
labels and state precedence in \cref{eq:reducer}.

This representation separates three operations that are often collapsed into
one prompt. \emph{Authoring} defines what must be true and which facts are
required. \emph{Extraction} populates those facts from document evidence.
\emph{Execution} applies the fixed state algebra. The distinction is material
for missing information: failure to extract a required fact is represented as
\texttt{missing}, whereas a source-supported contradiction is
\texttt{violated}. Neither can be converted to \meets{} by fluent explanatory
text.

Rules are stored independently from task references and model outputs. A task
adapter instantiates subjects and expected values, while source hashes identify
the documents against which the rule was evaluated. This permits a rule to be
reviewed or revised without changing the source adapter, and it permits the
same executor to consume facts from PDF, CAD, or IFC extraction. The seven
family definitions used here are listed in \cref{app:suite}; only their PDF
instantiations are evaluated.

\begin{figure*}[t]
\centering
\resizebox{0.98\textwidth}{!}{\begin{tikzpicture}[node distance=6mm and 7mm]
  \node[data] (pdf) {PDF pages\\\scriptsize evaluated};
  \node[data, below=3mm of pdf] (cad) {CAD / IFC\\\scriptsize adapter interfaces};
  \node[method, right=of pdf, yshift=-4mm] (extract) {Typed extraction\\text spans + visual regions};
  \node[method, right=of extract] (graph) {Fact graph\\predicate, values, state, source};
  \node[method, right=of graph] (rules) {Deterministic rule\\four-state reduction};
  \node[output, right=of rules, yshift=5mm] (decision) {Decision\\Meets / Does Not Meet\\Missing / Uncertain};
  \node[output, right=of rules, yshift=-10mm] (trace) {Proof trace\\rule steps + source spans};
  \node[method, right=of decision, yshift=-5mm] (select) {Escalation\\confidence + completeness};
  \node[output, right=of select] (expert) {Expert review\\when escalated};
  \draw[flow] (pdf) -- (extract);
  \draw[flow,dashed] (cad) -- (extract);
  \draw[flow] (extract) -- (graph);
  \draw[flow] (graph) -- (rules);
  \draw[flow] (rules) -- (decision);
  \draw[flow] (rules) -- (trace);
  \draw[flow] (decision) -- (select);
  \draw[flow] (trace) -- (select);
  \draw[flow] (select) -- (expert);
\end{tikzpicture}}
\caption{Constraint-checking architecture. Solid paths are instantiated for
PDFs in the experiment; the dashed CAD/IFC path is an adapter contract. The
rule engine emits both a decision and an auditable trace. The benchmark lacks
region-level reference annotations, so the experiment evaluates decisions and
finding content, not trace localization accuracy.}
\label{fig:architecture}
\end{figure*}
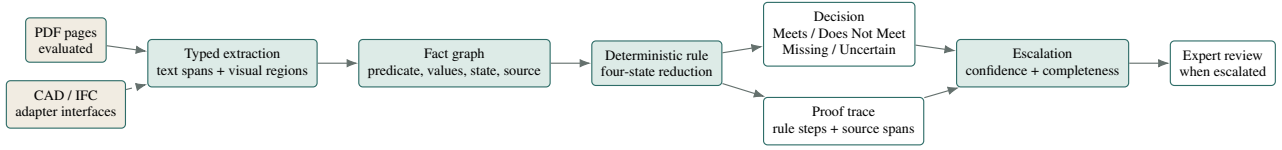

\subsection{PDF evidence allocation}

Layout-preserving text is extracted deterministically and indexed with BM25
\citep{robertson2009bm25}. The query is the task instruction with procedural
examples removed. Up to 12 page excerpts are retained, with document-balanced
retrieval for tasks containing at most four files. Images are rendered with a
maximum edge of 1,800 pixels.

The retriever indexes pages rather than arbitrary token windows so that every
hit has a stable page reference. Explicit page numbers in the task instruction
are inserted ahead of ranked candidates. For sets with at most four files,
each file contributes candidates before the global top-12 truncation; this
prevents a long specification or drawing set from suppressing every other
document. Each excerpt records its document ID, page, score, extraction method,
and content hash. These records are identical across the paired visual
conditions.

The \emph{page-RAG} policy uses up to four highest-ranked distinct page
overviews. The \emph{region-RAG} policy uses the same ordered text excerpts,
one overview of the highest-ranked page, and three overlapping half-span crops
of that page's ink bounding box. The crops begin at offsets 0, .25, and .5
along the longer content axis and are rendered from a 3,000-pixel source. Both
policies have a four-image cap; short documents can leave page-RAG slots
unused. A sensitivity analysis restricts the comparison to tasks where both
policies attach four images.

The crop axis is selected from the rendered page's content aspect ratio, not
from a task-specific annotation. Overlap avoids splitting a leader, note, or
detail exactly at a crop boundary. No expert reference, defect region, or
variant label enters retrieval or tiling. The intervention therefore tests a
generic allocation rule rather than an oracle crop around the answer.

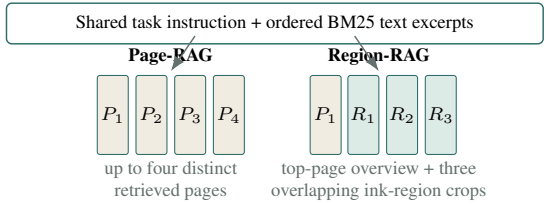
\begin{figure}[t]
\centering
\begin{tikzpicture}[x=.64cm,y=.55cm,font=\scriptsize]
  \node[block, minimum width=7.1cm] at (0,4.7)
    {Shared task instruction + ordered BM25 text excerpts};
  \node[font=\bfseries\scriptsize] at (-2.15,3.9) {Page-RAG};
  \node[font=\bfseries\scriptsize] at (2.15,3.9) {Region-RAG};
  \foreach \x/\n in {-3.35/1,-2.55/2,-1.75/3,-.95/4}{
    \draw[fill=CocoonSand,draw=CocoonTeal,rounded corners=1pt]
      (\x-0.32,1.55) rectangle (\x+0.32,3.45);
    \node at (\x,2.5) {$P_{\n}$};
  }
  \draw[fill=CocoonSand,draw=CocoonTeal,rounded corners=1pt]
    (.73,1.55) rectangle (1.37,3.45);
  \node at (1.05,2.5) {$P_1$};
  \foreach \x/\n in {1.83/1,2.63/2,3.43/3}{
    \draw[fill=CocoonMint,draw=CocoonTeal,rounded corners=1pt]
      (\x-0.32,1.55) rectangle (\x+0.32,3.45);
    \node at (\x,2.5) {$R_{\n}$};
  }
  \node[align=center,text=CocoonGray] at (-2.15,.92)
    {up to four distinct\\retrieved pages};
  \node[align=center,text=CocoonGray] at (2.15,.92)
    {top-page overview + three\\overlapping ink-region crops};
  \draw[flow] (-1.65,4.4) -- (-2.15,3.55);
  \draw[flow] (1.65,4.4) -- (2.15,3.55);
\end{tikzpicture}
\caption{Controlled evidence-allocation intervention. Text evidence and
response logic are fixed; only the four visual attachment slots differ.}
\label{fig:evidence-allocation}
\end{figure}

\subsection{Typed output, trace, and escalation}

The model returns the public task's typed records. A field-strict parser
rejects missing, extra, null, or empty fields. The deterministic rule layer
then maps records to \cref{eq:reducer}; it does not ask a second model to revise
the answer. A proof-trace object records the rule, ordered fact states,
confidence, and source references. Supported verified or violated facts fail
validation when no source is attached. The present task outputs do not contain
finding-level source references, so trace generation is an implemented
contract rather than an evaluated outcome.

Formally, the trace for rule $r$ is
$\pi_r=(r,d_r,\{(f_j,s_j,c_j,Z_j)\}_{j\in R_r})$. It is complete only when
every required fact is present, every fact has a source, and no state is
\texttt{uncertain}. Cross-document facts may cite multiple source references;
for example, a specification--drawing comparison can attach one clause span
and one drawing region to the same observed conflict. The trace therefore
preserves the evidence used by the reducer rather than a free-form rationale
generated after the decision.

The escalation interface flags uncertain decisions, missing information,
facts below a confidence threshold, and facts without sources. In the repeated
experiment, the available empirical confidence is the class frequency across
five calls. We evaluate whether a threshold of .8 lowers retained risk rather
than assuming that repeated agreement is calibrated.

Escalation is kept separate from the rule result. With threshold $\tau$, the
default action is
\begin{equation}
a_r=\mathbb{I}\!\left[d_r\in\{\mathrm{MI},\mathrm{U}\}
\;\lor\;\min_{j\in R_r}c_j<\tau
\;\lor\;\exists j:|Z_j|=0\right].
\label{eq:escalation}
\end{equation}
Thus a low-confidence supported violation remains \notmeets{} but is also
queued for review; escalation never rewrites it as a pass. The rule result,
trace completeness, and review action can consequently be audited as three
distinct outputs.

\section{Evaluation}
\label{sec:evaluation}

\subsection{Expert-reference constraint suite}

We include all \SuiteTasks{} tasks in seven \dataset{} families that require a
compliance, consistency, or defect judgment: reference resolution, technical
detail review, title accuracy, callout accuracy, sheet-index consistency,
specification--drawing synchronization, and submittal review. Retrieval-only
navigation and tracing tasks are excluded before analysis. The resulting suite
contains \SuiteProjects{} source projects and 160 task-level references. The
dataset card describes the annotations as expert-generated; no new labels or
inter-annotator agreement measurements are introduced here.

The suite is organized by source project, not by page or task variant. Its
repeated-test, development, and validation partitions contain 40, 74, and 46
tasks from 6, 18, and 5 disjoint projects. Across all partitions, the reference
decisions are 126 \notmeets{}, 26 \meets{}, and eight \missing{}. The
\uncertain{} class is available to the checker when an output is indeterminate;
it is not assigned to an expert reference. Each task is paired with a
machine-readable family rule specifying the relation and subject fields to be
checked. The complete family definitions appear in \cref{app:suite}.

Each reference is normalized to one of four decisions. A nonempty defect list
is \notmeets{} and an empty list is \meets{}. For submittals, any
\texttt{NOT\_MET} item is \notmeets; otherwise any \texttt{CANNOT\_VERIFY}
item is \missing; the remainder is \meets. Invalid or semantically
indeterminate predictions are \uncertain. The complete reference inventory,
source hashes, and family rule definitions are machine-readable.

Inclusion is registry-driven: every task in an included split and one of the
seven judgment families enters the suite. Tasks that only ask a system to find
a page or enumerate references are outside the target construct and are
excluded by family before comparative scoring. Several tasks are controlled
variants of the same documents; all such variants retain a common project ID
and are kept in the same split. This is why project, rather than task or model
query, is the unit used for resampling and exact tests.

Across the full suite, 146 references come from task-level ground-truth files,
12 from machine-readable reference lists embedded in the release, and two from
explicit clean-case sentinels. The build script records the source type and a
content hash for every normalized reference. This provenance makes the label
transformation auditable while preserving the source annotations as released.
It does not add an independent expert-adjudication layer.

\subsection{Systems and experimental blocks}

The repeated test uses four fixed deployments: Gemini 3.1 Pro Preview,
Gemini 3.6 Flash, GPT-5.6 Sol, and GPT-5.6 Luna. It contains all
\CoreTasks{} constraint tasks from the six-project held-out split. Each system
receives page-RAG, region-RAG, and a fixed family policy that selects one of
the two packets, with five calls per task and condition. This yields
\CoreTrials{} trials. The family policy is a routing ablation, not a learned or
uncertainty-adaptive router.

The panel contains two higher-capability and two efficient deployments. It is
used as a balanced intervention panel rather than a leaderboard: the primary
estimand gives each system equal weight, while system-specific effects remain
secondary. The five repetitions are fresh queries to the same fixed deployment
and packet; they measure response variability under the deployed inference
route.

The breadth extension was frozen before its model queries. It uses the
remaining \BreadthTasks{} tasks from \BreadthProjects{} disjoint development
and validation projects, the two Gemini systems, the two fixed evidence
policies, and one call per cell (\BreadthTrials{} trials). Because this block
was designed after inspecting the repeated test, it is a breadth extension,
not an independent confirmation. Results from the two blocks are reported
separately.

\subsection{Experimental controls}

Evidence packets are materialized before inference. Both fixed policies use
the unchanged task instruction, the same layout-preserving text extraction,
the same BM25 ranking, at most 12 text excerpts, the same typed response
contract, and the same model deployment. Only the allocation of up to four
image attachments changes. Filenames that directly expose a controlled task
variant are replaced by neutral document identifiers. Each packet is hashed
from its instruction, ordered excerpts, and image content; trial order is
randomized with a recorded seed. The response budget is 8,192 tokens.

A record-level pairing audit found no differences in instruction hash, ordered
text evidence, or required output fields across the 800 page/region trial
pairs in the repeated test and 240 paired cells in the breadth extension. This
check verifies the intended control in the archived packets rather than
assuming it from configuration alone.

The comparison is paired at the task--system level. A model first emits the
public task's structured records; the strict parser and deterministic reducer
then produce the four-way decision and typed finding set. Consequently,
page-RAG and region-RAG differ in visual evidence allocation, not in prompts,
text retrieval, output parsing, or decision logic. The family policy selects
one already-materialized packet by task family and introduces no additional
model call.

\subsection{Outcomes and inference}

The primary endpoint is four-way decision accuracy. Repetitions are first
averaged within task--system--condition cells. Tasks are then averaged within
project--family, projects within family, and the seven family means equally:
\begin{equation}
\bar Y_{mh}=\frac{1}{7}\sum_f\frac{1}{|P_f|}\sum_{p\in P_f}
\frac{1}{|I_{pf}|}\sum_{i\in I_{pf}}Y_{ipfmh}.
\label{eq:standardization}
\end{equation}
This prevents projects with many variants and large task families from
dominating the estimate.

Secondary outcomes are false-pass rate (predicting \meets{} for a
\notmeets{} reference), false-fail rate, exact task correctness, and typed
finding precision, recall, and F1. Findings are matched one-to-one: defect
matches require at least two distinct reference phrases; submittal matches
also require exact status and clause. Calibration uses the empirical
five-call class distribution, multiclass Brier score, and five-bin expected
calibration error. Selective risk is one minus accuracy among cells retained
at confidence at least .8.

False-pass rates are computed only over \notmeets{} references; false-fail
rates are computed only over \meets{} references and count an asserted
\notmeets{} decision. Typed finding scores use maximum-cardinality bipartite
matching so one broad prediction cannot receive credit for several reference
defects. Exact task correctness requires no unmatched prediction, no unmatched
reference, and no invalid record. This separates a correct four-way triage
decision from a complete professional finding set.

The primary contrast is region-RAG minus page-RAG, averaged over the fixed
systems in each block. We decompose this fixed-weight estimand into additive
source-project contributions. Percentile intervals resample those independent
contributions with replacement (10,000 draws), and exact two-sided sign-flip
tests enumerate their signs. System-specific contrasts are secondary and Holm
adjusted \citep{holm1979simple}.

These two uncertainty summaries answer different finite-sample questions:
the bootstrap describes variation under project resampling, whereas the exact
test evaluates sign assignments to the observed contributions. With few
clusters they need not cross zero at the same threshold, so we report both and
do not infer significance from a percentile interval alone.

The contribution bootstrap keeps the family, project, and task weights in
\cref{eq:standardization} fixed while resampling the independent project
clusters. The exact test enumerates every sign assignment to the same
contributions. Repetitions reduce within-cell noise but do not increase the
independent-project count. We report the repeated and breadth blocks
separately because they answer different questions: repeated stability across
four systems versus coverage across more projects for two systems.

\section{Results}
\label{sec:results}

\begin{table*}[t]
\centering
\caption{Repeated-test performance against expert references. Decision values
are project--family standardized percentages. FP is the false-pass rate among
\notmeets{} references; finding F1 uses one-to-one typed matching. The family
policy is a fixed routing ablation.}
\label{tab:main-results}
\small
\begin{tabular}{@{}lrrrrrr@{}}
\toprule
System & Page-RAG & Region-RAG & Family policy & $\Delta$ & Region FP & Region finding F1 \\
\midrule
Gemini 3.1 Pro & 38.6 & 56.1 & 50.5 & +17.5 & 37.9 & 12.6 \\
Gemini 3.6 Flash & 49.0 & 51.8 & 54.0 & +2.9 & 43.7 & 16.2 \\
GPT-5.6 Sol & 38.6 & 46.9 & 46.4 & +8.3 & 43.3 & 4.2 \\
GPT-5.6 Luna & 35.2 & 49.0 & 46.6 & +13.8 & 39.4 & 2.9 \\
\midrule
Panel mean & 40.3 & \textbf{51.0} & 49.4 & \textbf{+10.6} & 41.1 & 9.0 \\
\bottomrule
\end{tabular}
\end{table*}

\subsection{Repeated four-system test}

Region-RAG improves pooled standardized decision accuracy from 40.3\% to
51.0\%, a gain of \CoreDelta{}~\pp{} (95\% CI
\CoreCILow{}--\CoreCIHigh{}; exact $p=\CoreP$). Every source project's
weighted contribution to the pooled contrast is positive. The four
system-specific estimates are also positive, ranging from 2.9 to 17.5~\pp{},
but none is individually significant after Holm correction. The pooled panel
is the primary contrast of this reanalysis, not a claim about any one
deployment.

\Cref{tab:main-results} shows the associated safety and localization outcomes.
The mean standardized false-pass rate falls from 51.6\% under page-RAG to
41.1\% under region-RAG, and the false-fail rate falls from 35.2\% to 29.4\%.
These are meaningful reductions but leave absolute error rates too high for
unsupervised use. Exact task correctness rises only from 15.9\% to 20.5\%, and
mean typed finding F1 from 5.2\% to 9.0\%. Region-focused evidence therefore
improves coarse triage more than complete defect localization.

The fixed family policy reaches 49.4\% decision accuracy, below always using
region-RAG (51.0\%). Its false-pass rate is slightly lower, but its false-fail
rate is higher. Family identity alone is not sufficient routing information;
an adaptive policy would need evidence-quality signals that generalize across
projects.

\subsection{Heterogeneity and paired error audit}

The descriptive family contrasts are positive in all seven families but vary
substantially. Region-minus-page accuracy is 26.1~\pp{} for cross-reference
resolution, 16.2 for technical-detail review, 13.3 for sheet-index checks, and
10.0 for title accuracy. It is 4.6 for submittal review, 2.5 for
specification--drawing synchronization, and 1.7 for note-callout accuracy.
At the task level, 25 of 40 pooled contrasts are positive, eight are zero, and
seven are negative. These counts are descriptive because variants from the
same source project are not independent.

The largest positive and negative task contrasts illustrate both directions.
On a local reference-resolution
task containing an incorrect \texttt{3/A355} citation, region-RAG produces the
correct decision in 18 of 20 system--repetition trials, compared with 9 of 20
for page-RAG. Yet only seven region-RAG outputs match the typed defect,
showing that decision correctness can overstate report completeness. On a
cross-document door and glazing task with three specification conflicts,
page-RAG is correct in 7 of 20 trials and region-RAG in only 1 of 20. The
focused packet can expose small local text while withholding visual breadth
needed for a multi-document comparison.

\begin{figure}[t]
  \centering
  \IfFileExists{figures/contrast_forest.pdf}{%
    \includegraphics[width=\columnwidth]{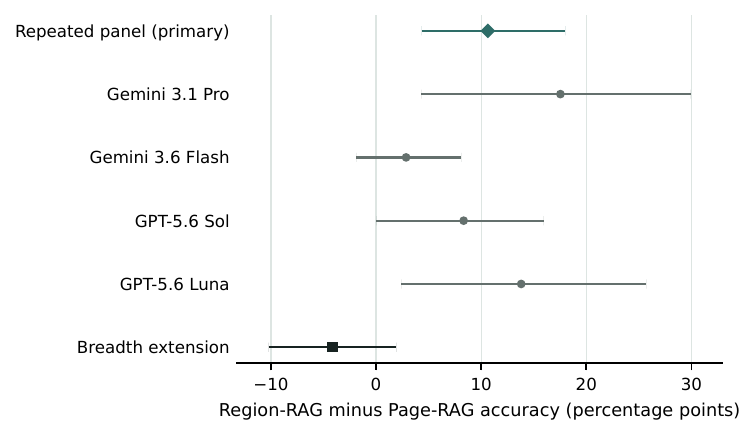}%
  }{\fbox{\parbox[c][1.7in][c]{.94\columnwidth}{\centering Result figure generated after analysis.}}}
  \caption{Region-RAG minus page-RAG decision accuracy. Intervals are
  source-project cluster bootstraps. The repeated-test panel is the primary
  contrast; system rows and the breadth extension are secondary.}
  \label{fig:contrasts}
\end{figure}

\subsection{Disjoint project breadth extension}

Across \BreadthTasks{} additional tasks and \BreadthProjects{} disjoint
projects, the two-system pooled contrast is \BreadthDelta{}~\pp{} (95\% CI
\BreadthCILow{}--\BreadthCIHigh{}; exact $p=\BreadthP$). The equal-four-image
sensitivity is $-9.7$~\pp{} over 86 tasks from 16 projects (95\% CI
$-19.2$--$0.0$; exact $p=.082$). Development and validation estimates are
$-3.1$ and $-10.4$~\pp{}, respectively. Both system-specific breadth
contrasts are negative ($-4.8$ and $-3.4$~\pp{}; Holm-adjusted $p=.806$).
Five project contributions are positive, 11 negative, and seven zero.
\Cref{app:additional-results} gives the full sensitivities.

Average standardized decision accuracy in this block is 55.5\% for page-RAG
and 51.3\% for region-RAG. Region-RAG reduces the mean false-fail rate from
41.5\% to 27.9\% but increases the false-pass rate from 43.0\% to 46.7\%.
Finding F1 also falls from 18.0\% to 16.1\%. This block tests project breadth
with one call per cell; it does not estimate repeated-run uncertainty. Its
opposite point estimate shows that the repeated six-project gain is not stable
under a fixed region policy across the remaining projects.

\subsection{Calibration and selective risk}

Across systems and evidence policies, five-call majority confidence is high
(mean .86--.93) even when decisions are wrong. Expected calibration error
ranges from .36 to .67. Escalating cells below .8 confidence retains
75--93\% coverage, yet retained decision risk remains 43--80\%. Repetition is
useful for measuring instability, but agreement frequency is not a validated
confidence score for this task. A deployable escalation policy requires
confidence tied to evidence completeness or an independently calibrated
correctness model.

\section{Discussion}

\paragraph{What the evidence contrast identifies.}
The intervention keeps the instruction, text retrieval, output schema, and
deployment fixed while changing how a maximum of four image attachments are
allocated. Region-focused evidence improves the repeated six-project panel,
whereas the 23-project breadth estimate favors page coverage and includes zero.
The intervention therefore identifies a resolution--breadth tradeoff, not a
dominant packet. Page breadth can be essential for cross-document checks, and
short documents allow region-RAG to use attachment slots that page-RAG cannot
fill. The equal-four-image sensitivity removes that slot-count difference and
still favors page-RAG, although its exact project test remains inconclusive.

\paragraph{The estimand is architecture-level.}
The pooled contrast gives equal weight to four fixed deployments after pairing
conditions within task and system. It therefore asks whether changing the
evidence packet shifts decisions across this panel; it is not an average model
score or a system ranking. All four system estimates are positive, but
their magnitudes differ and the six-project system-specific tests are coarse.
Both systems in the broader block have negative point estimates. The combined
result does not justify a model-independent recommendation for either fixed
packet and instead motivates validation of a routing policy on new projects.

\paragraph{Decision and finding accuracy differ.}
A ten-point decision gain can coexist with very low exact finding accuracy.
The four-way label asks whether a constraint fails; a professional report must
also identify the right object, location, clause, and explanation. This gap is
why the architecture separates deterministic decision reduction from proof
traces. Future datasets should annotate source regions and cross-document
links so that trace correctness can be measured directly rather than inferred
from answer text.

\paragraph{Implications for hybrid systems.}
The family-policy ablation does not outperform a fixed region policy, and
replicate agreement does not produce reliable selective risk. Useful routing
signals must therefore come from the evidence state: retrieval coverage,
unresolved entities, missing source links, conflicting observations, or
calibrated fact confidence. The typed engine exposes these variables and can
escalate before a pass is issued. The present experiment does not evaluate a
learned router or establish a production-ready confidence model.

\paragraph{Passes require evidence completeness.}
The four-state output prevents two operationally different cases from sharing
a pass label. \missing{} means a required source or value was absent;
\uncertain{} means the available evidence did not support a stable resolution.
Only \meets{} requires every rule fact to be verified. This distinction is
especially important when false passes remain common: an extractor that does
not observe a defect has not thereby produced evidence that the constraint is
satisfied. In a review interface, missing and uncertain cases can be routed to
different queues while a trace-complete \notmeets{} case can present the cited
conflict directly to an expert.

The benchmark outputs do not expose fact-level source links, so this
operational policy is an architectural contract rather than a measured safety
improvement. A deployment study would need experts to adjudicate both the
decision and every cited region, then estimate risk at each escalation
threshold on genuinely new projects.

\paragraph{Evidence budgets should follow constraint structure.}
Region-RAG spends three attachment slots increasing spatial resolution on the
highest-ranked page. This is well matched to titles, details, leaders, and
local references, but it narrows visual page coverage. A production allocator
should therefore condition on the rule's evidence graph: local geometric
predicates need resolution, whereas index checks and specification comparisons
need breadth across pages and files. The fixed family policy is only a coarse
version of this decision and leaves no mechanism to detect an atypical task
within a family.

\paragraph{Transfer to other professional documents.}
The source contract and four-state reducer do not depend on construction
terminology. They can represent a clause linked to an exhibit, a maintenance
requirement linked to an equipment schedule, or a clinical criterion linked
to a report region. Transfer would still require domain-authored rules,
source adapters, and project-level validation. The construction results alone
do not establish performance in those domains.

\section{Limitations}

The expert references are inherited from the public release. Their provenance
is documented as expert-generated, but independent relabeling and
inter-annotator agreement are unavailable. The strict phrase matcher can miss
valid paraphrases and should be replaced by adjudicated labels in a deployment
study. The label distribution is also strongly weighted toward
\notmeets{} cases; project--family standardization controls task volume, not
class prevalence.

The repeated test is a reference-based analysis of an existing experiment and
has only six independent projects. The 23-project breadth extension was frozen
before its model queries, but it was designed after the repeated result was
known; it is not an independent confirmation. It also has one call per cell
and only two systems. The source tasks are public, so training-data exposure
cannot be excluded. Inference concerns these project distributions and the
named hosted deployments at the evaluation date, not all construction
documents or future checkpoints.

With six clusters, the repeated-test bootstrap interval is necessarily coarse
and cannot create project diversity absent from the data. The exact project
test is reported alongside it, and no system-specific repeated-test contrast
survives multiplicity correction.

Only PDF inputs are evaluated. CAD and IFC are represented by tested source
and adapter contracts, not empirical modality results. The benchmark does not
provide region-level proof annotations, so finding-level trace correctness is
not measured. Finally, the image policies share a four-attachment cap rather
than exact pixel equality, and page-RAG can use fewer than four images on short
documents; the restricted sensitivity quantifies this design boundary.

The experiment isolates a complete packet-level intervention, not every
mechanism inside it. Region-RAG changes page coverage, local resolution, and
the number of occupied attachment slots on some tasks. The equal-four-image
sensitivity removes the last difference but does not equalize rendered pixels
or visual tokenization. Likewise, the fixed family policy is not a substitute
for a trained evidence router, and the repeated-agreement analysis does not
evaluate fact-level confidence from the typed engine.

\section{Conclusion}

Evidence allocation materially changes constraint decisions in professional
documents. Across a repeated four-system test, concentrating a bounded visual
packet on one retrieved page and its regions improves expert-reference
decision accuracy by \CoreDelta{}~\pp{}, with positive contributions from all
six source projects. Across 23 disjoint breadth projects, however, the estimate
is \BreadthDelta{}~\pp{} and both tested systems favor page-RAG. A fixed
region-focused packet is therefore not a general solution to large-format
document review. The practical direction is a hybrid checker that routes
evidence according to the constraint, treats missing and uncertain facts
explicitly, and exposes source spans and proof completeness to experts. The
current systems remain unsuitable for autonomous approval: false passes are
frequent, exact localization is low, and repeated agreement is miscalibrated.

\section*{Impact Statement}
Construction-document review affects safety, accessibility, cost, and legal
compliance. The evaluated systems have high false-pass rates and low exact
finding localization; they are not suitable for autonomous approval. The
intended use of this work is to improve evidence presentation, auditing, and
expert triage while preserving professional responsibility.

\section*{Reproducibility Statement}
The artifact freezes task and rule registries, canonical responses, scoring
code, and figure sources. Recorded hashes regenerate every table and figure
from archived records. Tests cover source contracts and analysis cardinalities;
rebuilding the 160-task suite reproduces it byte for byte.

{\sloppy
\bibliography{references}
\bibliographystyle{icml2026}
}

\clearpage
\appendix
\onecolumn
\section{Constraint Suite}
\label{app:suite}

\begin{table}[h]
\centering
\caption{Machine-readable constraint suite by split. The repeated test uses
the 40-task split; the breadth extension uses the other two splits.}
\begin{tabular}{@{}lrrr@{}}
\toprule
Split & Tasks & Projects & Repetitions per fixed cell \\
\midrule
Repeated test & 40 & 6 & 5 \\
Development & 74 & 18 & 1 \\
Validation & 46 & 5 & 1 \\
\midrule
Total & 160 & 29 & -- \\
\bottomrule
\end{tabular}
\end{table}

The 160 references contain 126 \notmeets{}, 26 \meets{}, and eight
\missing{} decisions. The \uncertain{} class is reserved for indeterminate
predictions rather than assigned as a reference label. Reference inventories
come from 146 released \texttt{gt.json} files, 12 executable reference lists,
and two explicit clean-case sentinels. The artifact stores both source hashes
and the precedence rule used when these release components differ.

\begin{table}[h]
\centering
\caption{Family-level rule represented in the suite.}
\small
\begin{tabularx}{\textwidth}{@{}L{.20\textwidth}Y@{}}
\toprule
Family & Constraint \\
\midrule
Reference resolution & Target sheet and view exist; content is relevant when determinable. \\
Technical detail & The reviewed assembly has no in-scope technical conflict. \\
Title accuracy & View title agrees with view type, orientation, and subject. \\
Callout accuracy & Leader-note text agrees with the element at its endpoint. \\
Sheet index & Index entries and title blocks correspond in number and title. \\
Spec.--drawing sync & In-scope requirements do not conflict across documents. \\
Submittal review & Submitted evidence satisfies each in-scope specification requirement. \\
\bottomrule
\end{tabularx}
\end{table}

\section{Reference Normalization and Matching}
\label{app:scoring}

Predictions are parsed into the exact public task fields. For non-submittal
families, explicit no-issue records are not findings; other non-abstaining
records are findings. For submittals, \texttt{NOT\_MET},
\texttt{CANNOT\_VERIFY}, and \texttt{MET\_WITH\_NOTE} records remain typed
findings. Decision normalization follows the precedence stated in
\cref{sec:evaluation}.

Candidate matches are constructed after Unicode, case, whitespace, and
punctuation normalization. A defect prediction must contain at least
$\min(2,k)$ distinct phrases from a $k$-phrase reference. A nonempty canonical
replacement can supply one phrase. Submittal matches require exact normalized
status and specification clause in addition to two reference phrases. Maximum
cardinality bipartite matching enforces one-to-one assignment; unmatched
predictions are false positives and unmatched references false negatives.

This matcher is intentionally stricter than substring matching and transparent
enough to audit, but it is not a substitute for independent expert
adjudication. The artifact includes trial-level decisions, match counts, and
reference sources so alternative matching rules can be evaluated.

\section{Execution and Statistical Protocol}
\label{app:protocol}

Evidence packets are materialized deterministically before model calls and
identified by hashes of the instruction, ordered text excerpts, and images.
The public task instruction is retained verbatim. Filenames that reveal
controlled variant status are replaced with neutral document identifiers.
Both fixed policies use the same response contract. Trial order is randomized
by a recorded seed; repetitions index fresh calls without exposing or
controlling the inference sampler's internal seed.

For bootstrap inference, source projects are sampled with replacement and all
their task--family contributions are carried into the standardized mean.
Percentile bounds use 10,000 resamples. Exact tests compute each project's
weighted contribution to the paired contrast, enumerate all sign assignments,
and report the fraction whose absolute statistic is at least the observed
value. This yields $2^6$ assignments in the repeated test and $2^{23}$ in the
breadth extension. Calls and task variants are never treated as independent
projects.

\section{Additional Results}
\label{app:additional-results}

\begin{table}[h]
\centering
\caption{Breadth-extension sensitivities. $\Delta$ is Region-RAG minus
Page-RAG standardized decision accuracy in percentage points.}
\label{tab:breadth-sensitivity}
\small
\begin{tabular}{@{}lrrrrr@{}}
\toprule
Scope & Tasks & Projects & $\Delta$ & 95\% CI & $p$ \\
\midrule
All extension tasks & 120 & 23 & -4.1 & [-10.2, 1.9] & .209 \\
Both packets use four images & 86 & 16 & -9.7 & [-19.2, 0.0] & .082 \\
Development split & 74 & 18 & -3.1 & [-10.8, 4.5] & .450 \\
Validation split & 46 & 5 & -10.4 & [-18.5, -1.7] & .125 \\
\bottomrule
\end{tabular}
\end{table}

The repeated-test system contrasts are secondary. Gemini 3.1 Pro, Gemini 3.6
Flash, GPT-5.6 Sol, and GPT-5.6 Luna have region-minus-page estimates of 17.5,
2.9, 8.3, and 13.8~\pp{}, respectively. Their exact six-project $p$-values are
.125, .500, .250, and .125; all Holm-adjusted values are .500. The data support
the pooled repeated estimate, while the system-specific tests do
not distinguish the four effects from zero at the project level.

\section{Typed Rule and Proof-Trace Contract}
\label{app:trace}

Each fact contains \texttt{fact\_id}, \texttt{predicate}, \texttt{subject},
\texttt{expected}, \texttt{observed}, \texttt{state}, \texttt{confidence}, and
one or more source references. PDF references require a one-indexed page and
optional normalized bounding box. CAD and IFC references require an entity ID
and may include a property path. Verified and violated facts without evidence
fail validation.

The deterministic rule engine checks unique fact identities, inserts a
\texttt{missing} fact for any required fact not extracted, applies
\cref{eq:reducer}, and records ordered proof steps. The default escalation
policy flags uncertain decisions, missing information, minimum fact confidence
below .8, or any untraced fact. Unit tests cover state precedence, missing-fact
insertion, low-confidence escalation, normalized PDF regions, and CAD/IFC
entity requirements.

\end{document}